# Guidance of an Autonomous Underwater Vehicle in Special Situations

M. Eichhorn, Member, IEEE
Institute of Automation and Systems Engineering
Technical University of Ilmenau, P.O. Box 100565
Ilmenau, 98684, Germany
Mike.Eichhorn@TU-Ilmenau.de

*Abstract* - **This article describes a guidance system of the autonomous underwater vehicle "DeepC" [1] in special situations. A special situation occurs when one or more objects interfere with the planned route of a mission. The possible reactions are evasion or identification of the objects. The paper presents these two tasks in overview.**

**The special demands challenges of the underwater environment, computer parameters, sensors and the maneuverability of the vehicle are considered in the selection and development of the required strategies. Such challenges include the sea current, maneuver in the 3-D space and the limited perceptive faculty of the sonar.**

## I. INTRODUCTION

### A. Research project "DeepC"

An inter-disciplinary consortium including German companies and research institutes supported by the federal ministry of science and development is working on the development of an autonomous underwater vehicle [1], [2]. This vehicle should be able to undertake missions on its own and react independently in unexpected situations (e.g. objects on the planned course, component malfunction).

The vehicle is specified by the parameters in Table 1. These are realized by the following technical details:

- Separate payload- and vehicle hulls (Fig. 1)
- Redundant construction of all important systems and separate storage in the individual vehicle hulls
- Energy generation using Polymer Electrolyte Membrane PEM fuel cells
- Exchangeable payload container
- Specific software modules for the fulfillment of the required autonomy.

TABLE 1: PARAMETERS OF THE VEHICLE

| Parameters | Value |
|---|---|
| Limits of the vehicle | Length: 5800 mm |
| | Bright: 2300 mm |
| | Height: 1700 mm |
| Vehicle weight | 2,4 t |
| Depth of Diving | 4000 m |
| Cruise/Max speed | 4 kts / 6 kts |
| Endurance | 40 h |
| Vehicle payload | 300 kg |
| Equipment | TV-Camera, Sonar |

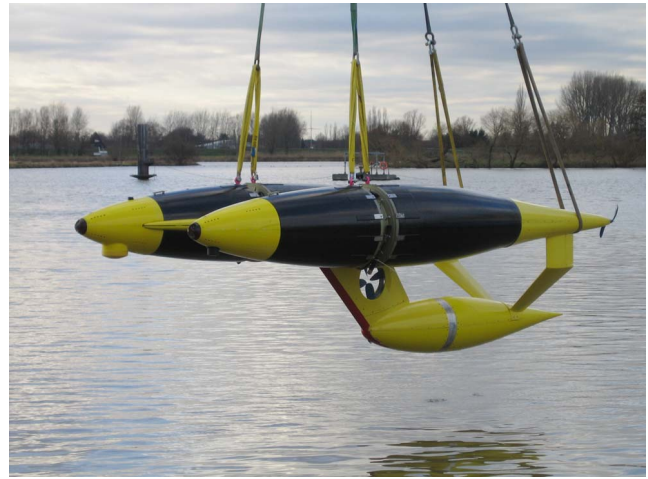

Fig. 1. The Autonomous Underwater Vehicle DeepC

Accurate vehicle positioning and maneuverability, while stationary and at low speeds, is achieved by thrusters placed in the vertical and the horizontal planes. At higher speeds, the vehicle is controlled by a combination of fin units, thrusters and differences in rotation speed of the main drive system.

### B. Control hierarchy of the vehicle

The control hierarchy of the vehicle is multi-level. This corresponds to the Brynes Rational Behavior Model (RBM) Architecture, which is applied in the AUV *NPS Phoenix* [3]. In this concept, the tasks are solved in the *Strategic level*, *Tactic level* as well as in the *Execution level*. This hierarchy concept is comparable with the working organization of a submarine.

Fig. 2 shows the major software modules of the vehicle control hierarchy. Normally the vehicle is working according to a mission plan. This plan consists of a sequence of complex maneuvers, which describe special tasks (Track, Meander, GPS Update) during the mission. The module *Missionplanhandler* converts the complex maneuvers to simple geometric shapes, the basic maneuvers. These basic maneuvers include positions and other geometric data, which are passed in a fixed order by the vehicle. Vehicle control requirements, such as setpoint -depth -path, -position and -speed are committed to the *Autopilot*. This mode is known as automatic control. The Module *Object Detection* generates objects in form of elliptical cylinders from the sonar basis data. If such objects tangents the path of the vehicle, other algorithms replace the automatic control.

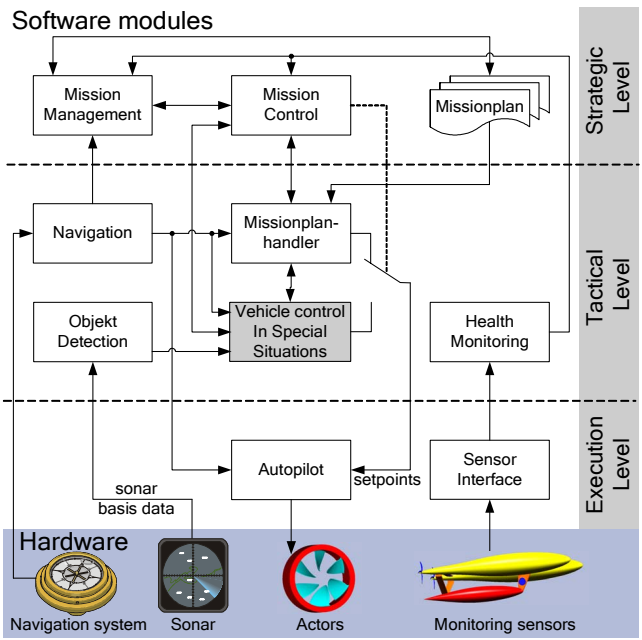

Fig. 2. Control structure of the vehicle

In accordance with the mission task these algorithms control the guidance of the vehicle: either for avoiding or for identifying the objects. These algorithms are the main components of the module *Vehicle control in Special Situations* (VCS) (depicted in Fig. 2 gray), which is introduced in this article. The coordination of the individual modules and/or the tasks, which can be accomplished take place via the *Mission Control* in the *Strategic level*. Thus the VCS module decides which maneuver is required when a special situation occurs. If an existing mission plan must be re-planned by the *Mission Management*, the *Mission Control* enables this first. Such a mission restructuring can take place in case of failure of hardware components, missing energy resources or if there is too large a navigation inaccuracy. The *Health Monitoring* determines the information about the vehicle state by a diagnosis of the hardware components.

*C. Stucture of the software module VCS*

The information about obstacles is given to the software module VCS in form of elliptical cylinders by the module *Object Detection* (see Fig. 3).

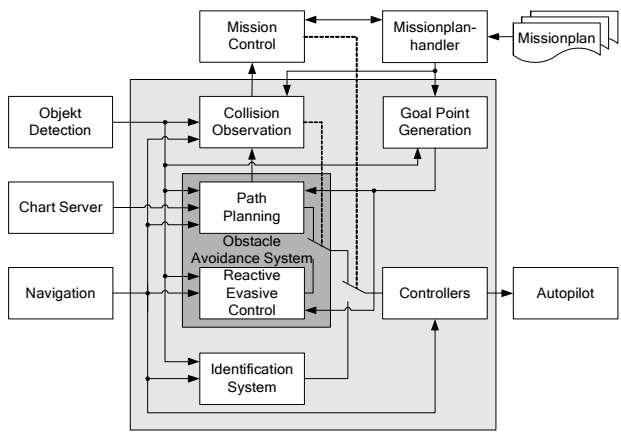

Fig. 3. Structure of the software module VCS

This form can be easily generated from the basis data of the sonar and characterizes many practical obstacle structures- with a small number of geometric variations: such as the major axis, the minor axis, the rotation angle of the ellipse and the height of the cylinder. The *Collision Observation* examines a collision possibility between the detected objects and the current basis maneuvers of the mission plan during the mission. If such a possibility is detected, this is announced to the module *Mission Control*. This module activates, dependent upon the current mission task, the *Obstacle Avoidance System* or the *Identification System*. The activated system takes over the guidance of the vehicle and hands the control information over to the sub-module *Controllers*. Such control information can either be defaults for target positions or the distance to an object. In the sub-module *Controllers* this control information is prepared in such a way that it is handed over to the autopilot in the form of desired values for course, depth and speed. In case of a special situation, the sub-module *Goal Point Generation* is creating a point of rendezvous with the target route of the mission plan, using the actual plan of mission and the detected objects, which is the goal point for the obstacle avoidance system.

## II. OBSTACLE AVOIDANCE SYSTEM

This section describes an overview of the obstacle avoidance system. Detailed information is presented in [4] and [5].

For the obstacle avoidance system a two level structure was favored (colored dark grey in Fig. 3). The upper level consists of *Path Planning*: a route avoiding the obstacles is generated here, using the existing information about the objects and the environment.

In case of unexpected or suddenly emerging obstacles, during navigation on the generated route, control will be passed to the lower level by the *Collision Observation*. The *Reactive Evasive Control* reacts on the located obstacles in the proximity of the sonar by reactive course commands. During the activation of the level *Reactive Evasive Control* the level *Path Planning* the route plan can be modified or recreated using the new object information.

Fig. 4 shows the routes generated using *Path Planning* and *Reactive Evasive Control* on an example of a possible obstacle scenario with fixed start and finish points. Here you can see the reactive ("local") reaction of the *Reactive Evasive Control* and the planning ("global") reaction of the *Path Planning* at point **A** very clearly.

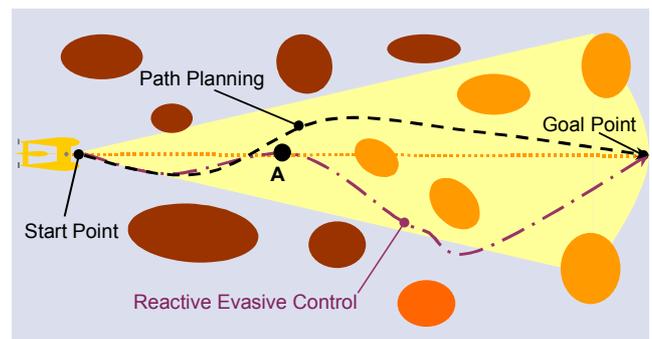

Fig. 4. Route of course by using the several levels of control

*A. Path Planning*

All information about the actual operation area is used for the path planning to generate a plan of the route. These are the actual information of the sonar, the stored data of obstacles of the passed mission and data of a digital sea map. Such a route plan is created based upon graphs. Therefore points (vertices) at the operation area are defiled which are passable by the vehicle. The passable connections between these points are recorded as edges in the graph. Every edge has a rating (costs, weight) which can be the length of the connection, the evolving costs for passing the connection or the time required for traversing the connection (see Fig. 5).

After generating such a graph a way (route plan), from the initial vertex (starting point) to the end vertex with the lowest total costs, will be created by search algorithms (Dijkstra- [6], A*- [7], D*- algorithm [8]). These algorithms analyze the graph to find a combination of edges, which connect the start and finish vertex with the lowest total costs.

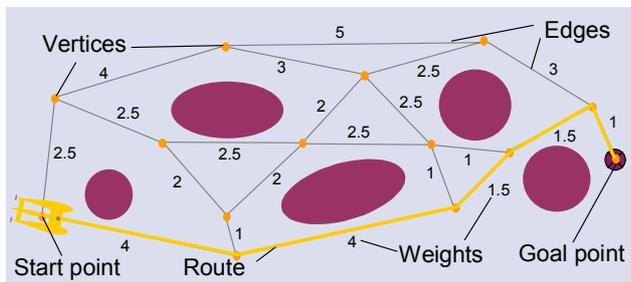

Fig. 5. Geometrical graph

In this project two methods are examined for the production of a geometrical graph (Visibility and Quadtree graph). The selection of these methods was based on the given obstacle form (elliptical cylinder) and the demand on computation time of the algorithms, under the given computational restrictions of the vehicle. The application of the methods in the 2D and 3D-world should be possible.

In the visibility graph, the corners of the obstacles form the vertices of the graph [9]. Obstacles without corners (circle, ellipse) are modeled by a polygon, which includes the obstacle completely. All possible lines between the vertices, which do not cut obstacles, become edges of the graph (see Fig. 6).

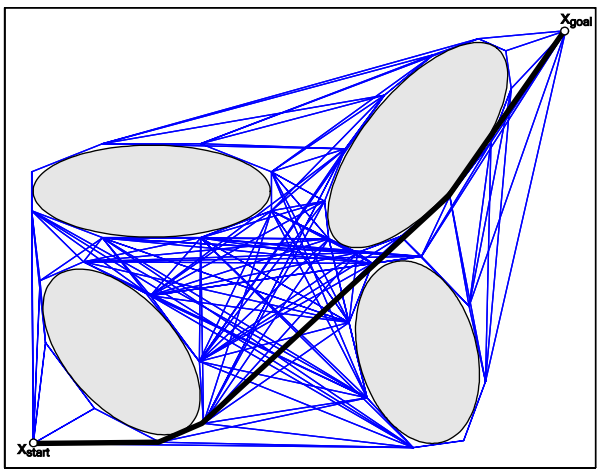

Fig. 6. Visibility graph

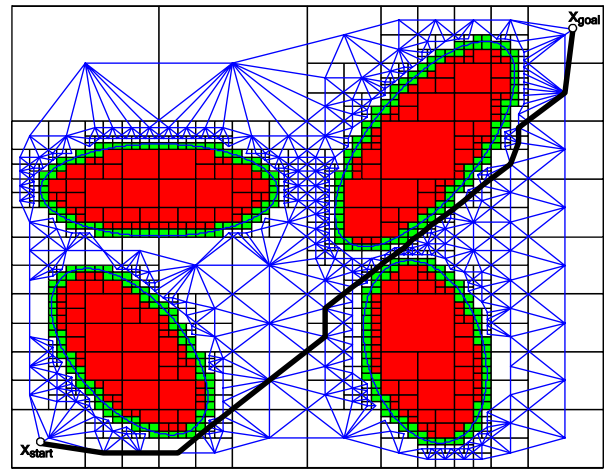

Fig. 7. Quadtree graph

In the Quadtree/Octree graph the area of operation is partitioned, into square (2D-plane) and/or cubic (3D-space) sectors, as a function of obstacles found in them, they can be "free" "occupied" or "partly occupied". Partly occupied sectors are recursively divided until a maximum division depth and/or a minimum sector size is reached. Sectors classified as "free" or "occupied" are not divided any more. The centers of all obstacle-free sectors form the vertices of the graph. The edges are the connections between neighboring obstacle-free sectors (see Fig. 7).

For the project the visibility graph was preferred. This is justified by its smaller computation cost and the form of the determined paths (compare Fig. 6 and Fig. 7).

*B. Reactive evasive control*

The reactive avoidance control must relieve the path planning in case of suddenly emerging obstacles and control the vehicle. Another mission case is created by input of „inexact" object data from the module *Object Detection*. Under "inexact" you can understand a stochastic variation of the dimensions and the position of the objects as a result of the limited range of the research sensor (e.g. Forward Looking Sonar) and the quality of detection dependent upon the distance. In such cases an evasive system, which has a high robustness to such variations, is required. This requirement is realized with the reactive evasive control. In this case the course commands (nominal course, nominal speed) for the autopilot system are only generated based on the actual position of the objects relative to the vehicle. A route planning, assuming a certain constancy of the data of the objects between the compilation und the overwork of the plan, was not leaded.

The idea of the method used in the *Reactive evasive control* consists of using trail lines, which create a way to the finish point from every position in the area of operation. The nominal course of the vehicle is determined upon creation of the course gradient (gradient G) to the actual vehicle position. In this case the vehicle can use the actual course line, respectively its gradient, in order to lead through the obstacles to the target point. The method of the reactive level is based on the works of [10]. In this case the lines of gradients are created by synthetic harmonic dipole-potentials (see Fig. 8).

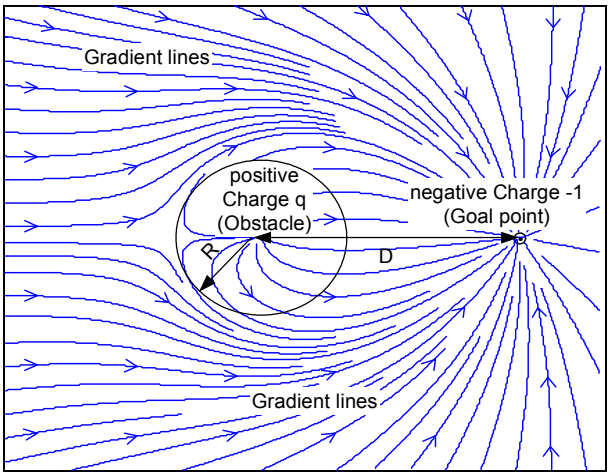

Fig. 8. Gradient lines of the electrical dipole

By using such dipole-potentials the negative characters of the synthetic potential fields can be compensated. Such characters are the trapping of the vehicle in local minima, the oscillating movements in small passages and the blockade of possible routes by the resulting revolting power field on the beginning of a passage [10]. For this method strong steering commands are necessary in order to keep the vehicle within the created line of gradients. The structure of the gradient lines in this area and on the lateral passing do not correspond with an optimal course route. Therefore a new method was developed which allows creation of gradients from a geometric construction.

In this case the operation area is divided into separate sectors (see Fig. 9). The classification of the sectors results dependent on the vehicle position $x_{veh}$ to the centre of obstacle $x_{obst}$ and to the target point.

If the condition ($|x_{veh} - x_{obst}| < R$) is discharged the vehicle is located inside the security circle. Sector 3 is active. If the view between the vehicle and the target point is blocked by the circle a direct route to the target point is not possible (sector 1). In this case the direction vector of the tangential line of $x_{veh}$ to the circle must be created dependence on the position of $x_{veh}$ to the symmetry axis. In case of a direct connection between the vehicle and the target point sector 2 is active. The resulting lines of gradients are present in Fig. 10.

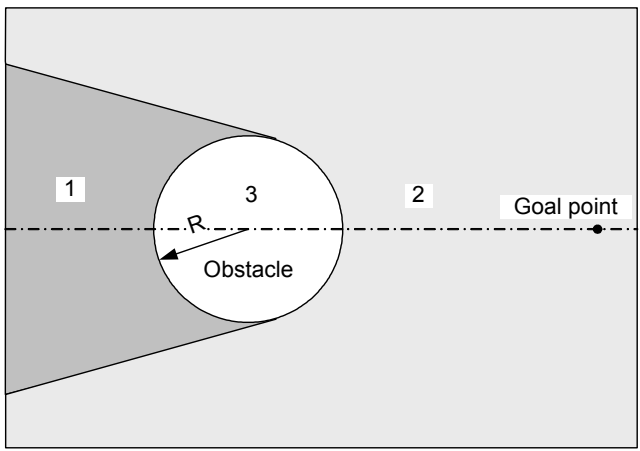

Fig. 9. Division of the area of operation into sectors

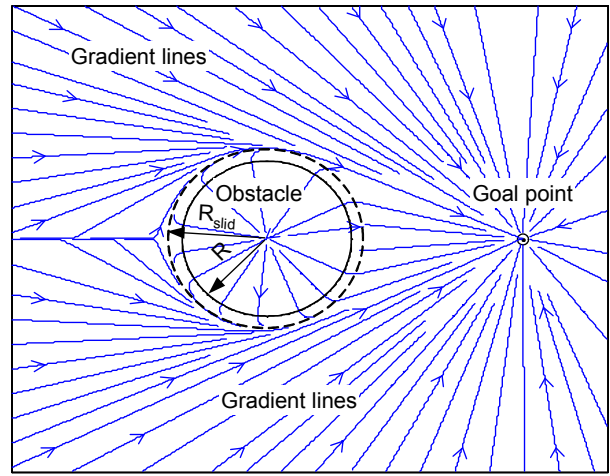

Fig. 10. Gradient lines by geometrical construction

A comparison of the generated routes between the new designed method of the geometric construction with the method of the harmonic dipole potentials shows Fig. 11.

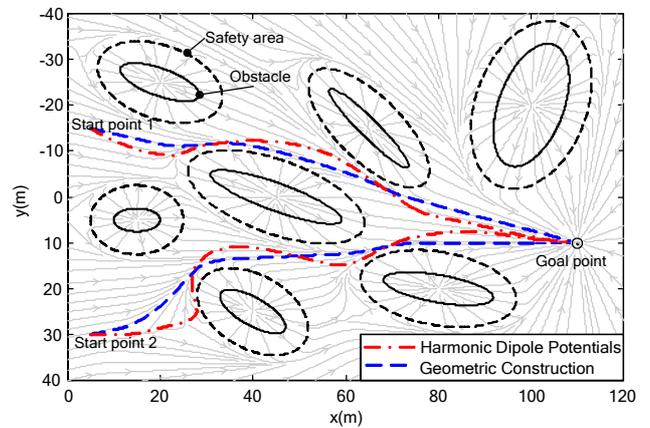

Fig. 11. Paths through the obstacle scenario

It could be shown that with the use of the geometrical construction method the target course change rate r possesses substantially smaller maximum values than by using the harmonic dipole potentials (see Fig. 12). This is an important criterion, since the applied vehicle cannot follow reference trajectory changes very fast, which causes deviations in the gradient lines and thus to possible collisions.

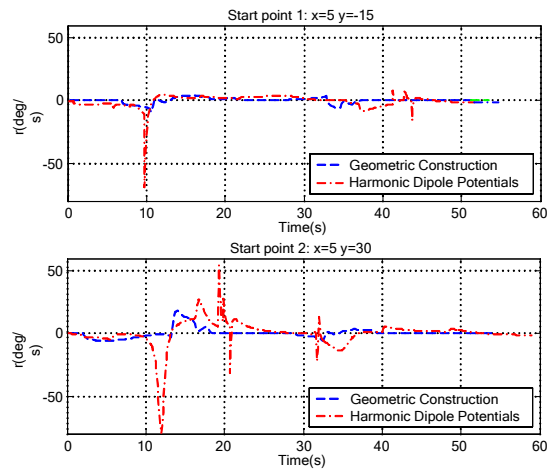

Fig. 12. Evaluation of the reference turning rate r

## III. IDENTIFICATION SYSTEM

Identification of objects can be one of the required tasks during a mission. It can include collection of visual or cartographical information or determination of a physical/chemical property. Therefore, the vehicle guidance handover to the sub-module *Identification System*. Here, the reference values for the underlying autopilot are determined using the object geometry and the requirements of the vehicle during identification (distance to object, orientation to object, moving velocity around the object). (see Fig. 13). Depending on the object size the operation mode can be divided into two: *aligning to an object* and *going around an object*. They will be introduced in the next sections.

After identification the goal point on the actual mission route is determined by the sub-module *Goal Point Generation* (see Fig. 3). The vehicle is then led to this goal point by the activated *Obstacle Avoidance System*.

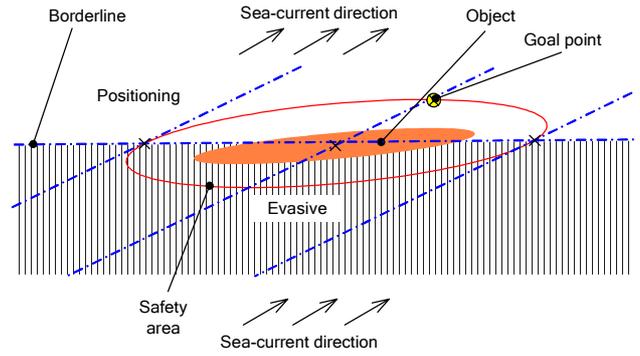

Fig. 14. Positioning to an Object in dependence of the sea-current

Fig. 15 shows the trajectory of an "Aligning to an Object" – Maneuver by a strong sea current. In the beginning of the maneuver, the vehicle avoided the object in Evasive Mode. As the vehicle drives over the borderline, the Positioning Mode is activated. The vehicle turns in the sea current at defined distance to the object.

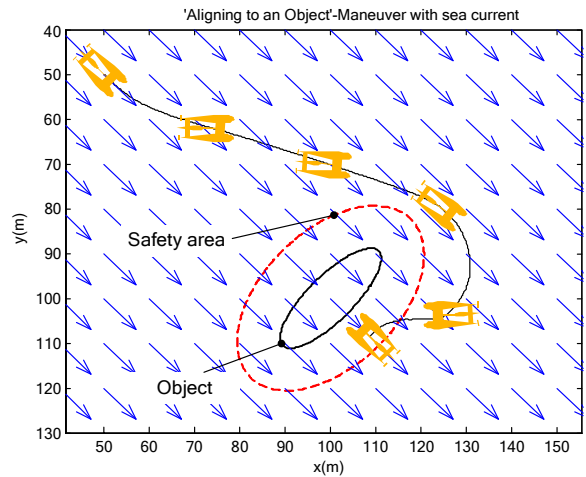

Fig. 15. Trajectory of an "Aligning to an Object" - Maneuver

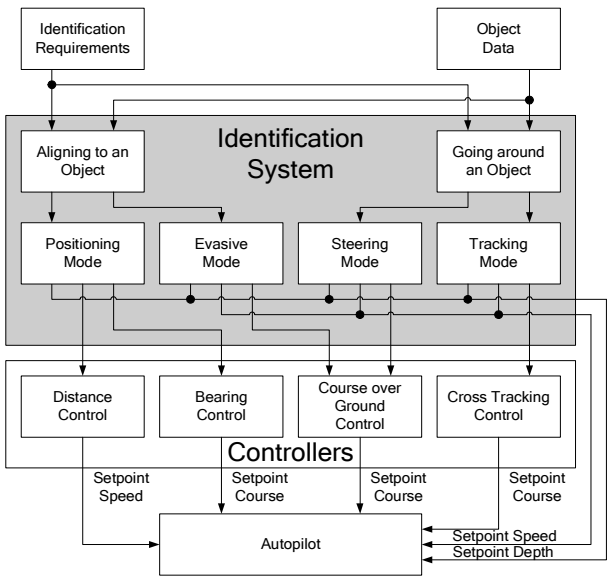

Fig. 13. Structure of the Identification System

### A. Aligning to an object

In order to be able to accomplish an identification, the vehicle must take a defined orientation to the object and remain like that during the identification procedure. This mode is called *aligning to an object*. If a strong current occurs in this mode, the vehicle must be led in the flow shade of the object, in order for it to take an accurate position and be able to hold it.

Hereby, areas around the object are defined considering the see current and safety areas, in which the vehicle must fulfill different of maneuver (see Fig. 14). If the vehicle is in the sea current in front of the object, it must avoid this object (Evasive AREA). For this maneuver the reactive evasive control algorithm, which was presented in the last section, will be used. To do this, the object was increased to a safety range and the goal point was chosen as the intersection point between the ray of the sea current direction with the ellipse of the safety area in the flow shade. If the vehicle lies in the flow shade the aligning to the object takes place with a defined distance (Positioning AREA).

### B. Going around an object

By more spacious objects, the vehicle must go around the object to be identified at a defined distance and orientation. This mode is called *going around an object*. If the vehicle is far away from the given target trajectory it switches to the Steering mode and is led on the shortest way to the target trajectory. Then the vehicle is switched to the Tracking mode to drive around the object in the given target distance.

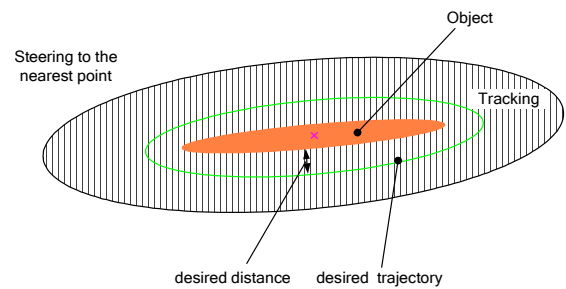

Fig. 16. Drive around an object

## IV. CONCLUSIONS

In this article a concept for the guidance of an underwater vehicle in special situations was presented. The special characteristics of the vehicle (existing computer performance and sensor technology, driving characteristic and maneuverability) formed the main criteria for the selection and development of the algorithms. The presented method for the geometrical construction of gradient lines unites the advantages of the method of the artificial harmonious dipole potentials with the demands on optimal path and control driving fashion. By using the graph methods for the determination of a route suggestion in the course planning level the demand on an energy-optimal driving fashion could be easily considered. In the identification mode the area of operations will be divided into separate areas into which special tasks of maneuver are to be solved depending on the object geometry and the sea-current. For this, special automatic controllers are used, which calculates reference values for the course, speed and depth for the underlying autopilot.

The evasive system was successfully tested in summer 2004 with the AUV MARIDAN 600 [11]. In spring 2005 began the shallow water tests on the goal platform "DeepC".

**Acknowledgments**

This work was a part of the research project "DeepC", which was promoted by the BMBF. (Promotion no: 03SX104E). I would like to dedicate this paper to the 65[th] birthday of Professor J. Wernstedt, who made it possible for me to work in this interesting project. I thank Uwe Möller and Hendrik Schelenz from ATLAS ELEKTRONIK GmbH Bremen for the good teamwork as well as Sandy McPherson from Allset Consultancy BV for the software-technical support and the helpful discussions.